\pdfoutput=1
\documentclass[11pt]{article}

\usepackage[preprint]{acl}

\usepackage{times}
\usepackage{latexsym}
\usepackage{datetime}
\usepackage{enumitem}
\usepackage{amsmath}
\usepackage{booktabs}
\usepackage{multirow}
\usepackage{subcaption}
\usepackage{caption}
\usepackage{listings}
\usepackage{float}
\usepackage{rotating}
\usepackage{makecell}
\usepackage{tikz}
\usepackage{ifthen}
\usepackage{etoolbox} % For conditional checks
\usepackage{fontawesome5}

\usepackage{lipsum}

\usepackage[T1]{fontenc}
\usepackage[utf8]{inputenc}

\usepackage{microtype}

\usepackage{inconsolata}

\usepackage{graphicx}
\usepackage{tabularx}

\title{Signs of Struggle:\\Spotting Cognitive Distortions across Language and Register}

\author{Abhishek Kuber, Enrico Liscio, Ruixuan Zhang,\\
{\bf Caroline Figueroa,} \and {\bf Pradeep K. Murukannaiah} \\
Delft University of Technology, the Netherlands \\
\texttt{abhi.kuber@gmail.com}\\ \texttt{\{e.liscio,r.zhang-2,c.figueroa,p.k.murukannaiah\}@tudelft.nl}}

\begin{document}
% \input{latex/title_page.tex}
% \clearpage

% \input{acknowledgements.tex}

\maketitle
\begin{abstract}
Rising mental health issues among youth have increased interest in automated approaches for detecting early signs of psychological distress in digital text. One key focus is the identification of \textit{cognitive distortions}, irrational thought patterns that have a role in aggravating mental distress. Early detection of these distortions may enable timely, low-cost interventions. While prior work has focused on English clinical data, we present the first in-depth study of cross-lingual and cross-register generalization of cognitive distortion detection, analyzing forum posts written by Dutch adolescents. 
Our findings show that while changes in language and writing style can significantly affect model performance, domain adaptation methods show the most promise.
\end{abstract}

\section{Introduction}
\label{sec: introduction}

Mental health disorders among adolescents are a growing global concern. According to the \citet{mentalillnesswho}, one in seven individuals aged 10-19 experiences a mental disorder, with depression, anxiety, and behavioural disorders being the most common.
This is particularly problematic in adolescence, where unaddressed conditions can have lasting effects into adulthood, highlighting the need for early, non-pharmacological interventions \cite{mentalillnesswho}.

Cognitive Behavioral Therapy (CBT) is a widely used treatment for mental health disorders \cite{BECK1970184,cbtgoldstandard,cbtfirstline}. It emphasizes that our interpretations of events -- not the events themselves -- determine how we feel. For instance, viewing a breakup as ``No one will ever love me again'', over time, may lead to social withdrawal and loneliness. These negative thought patterns, known as \textit{cognitive distortions}, are linked to conditions like depression and anxiety \cite{BECK1970184, PERSONS2023104338}. 
% Early detection of cognitive distortions can prevent long-lasting mental health effects.
By helping individuals recognize and reframe distorted thoughts, CBT can prevent long-lasting mental health effects. % (see Appendix~\ref{subsection:appendix_reframing}).

Despite rising awareness, many cases go undetected and untreated by conventional clinical approaches. An emerging trend is instead to analyze social media data on digital platforms, which captures authentic expressions of emotion and help-seeking behavior \cite{Chancellor2020-xe}.
%With 96\% of the adolescent population using the internet daily \cite{actyouth},
%Digital platforms have become key spaces where mental health struggles are voiced. 
One such platform is De Kindertelefoon\textsuperscript{\ref{kt-fn}}, where Dutch youth aged 8-18 can discuss issues such as sexuality, bullying, and emotional struggles on anonymous forums. The forums offer valuable insights into youth mental health, and provide a unique opportunity to explore automated techniques for supporting adolescent mental wellbeing.

On large-scale forums like De Kindertelefoon, manual review of every post is unfeasible, making automated detection of cognitive distortions a crucial first step.
However, prior work has focused mainly on English clinical data \cite{Shreevastava2021,sharmacognitivereframing, zhan2024largelanguagemodelscapable}, which differs from De Kindertelefoon data both in language and \textit{register} -- a shift from adult to adolescent writing that introduces added challenges.

To illustrate how the same topic can be expressed differently across registers, an adult might write, \textit{``I have felt lonely just about all my life\dots I really don't know who I am since I no longer am a hands on mother\dots I am lonely, confused and miserable''} (example taken from \citet{Shreevastava2021}). 
In contrast, a post on De Kindertelefoon (paraphrased into English) might say, 
\textit{``I have bad grades at school\dots who can I talk to about these things because I don't have any good friends that I can trust with this\dots Please help me because I can't do it anymore''}.
Both posts discuss the feeling of loneliness, but the adult’s post is reflective and provides context, while the adolescent’s post lacks elaboration. 
These differences showcase the challenge of building models that can generalize across registers.

We perform the first in-depth study of how computational methods for cognitive distortion detection generalize across both language and register.  
Our experiments range from prompting to supervised learning and domain adaptation, evaluating generalization across both language and register shifts. 
Our results show that, while multilingual models can generalize across languages, they often struggle with register changes such as writing style, and domain adaptation proves essential for improving performance. Overall, our work demonstrates that cognitive distortion detection can be adapted across languages and registers -- a critical first step towards making them more generalizable.

\section{Background and Related Work}

We review related works on automated CBT detection and domain adaptation techniques.

\subsection{Computational Approaches to CBT}

Early approaches to detecting cognitive distortions in text relied on linguistic features \cite{8031202}.
With the rise of transformers, supervised learning has gained traction. 
\citet{Shreevastava2021} compare semantic and syntactic feature types and show that combining Sentence-BERT embeddings with SVMs improves performance. \citet{jiang2024aienhancedcognitivebehavioraltherapy} frame distortion detection as a hierarchical classification task using a supervised model pretrained on knowledge graphs.

The surge of LLMs has shifted the attention to prompt-based approaches. \citet{Chen2023} introduce Diagnosis-of-Thought prompting, a Chain-of-Thought approach grounded in cognitive theory. \citet{Lim2024} propose ERD, combining extraction and debate across multiple LLMs. TeaBot \citep{Nazarova2023} uses GPT-3 for real-time distortion detection using CBT-inspired questions.

%Building on these strategies, we combine contextual embeddings with lexical features to improve classification performance. 
We compare prompting, instruction tuning, and supervised fine-tuning approaches to cognitive distortion detection. However, we show that they do not generalize across registers (Section~\ref{subsection:stage1_preliminary_methods}), highlighting the need for domain adaptation techniques.

\subsection{Domain Adaptation}
In NLP, a domain refers to a coherent corpus shaped by topic, style, or language. Domain adaptation tackles the challenge of applying models trained on one domain to another, facing performance drops due to such variations \cite{ramponi-plank-2020-neural}. Various strategies have been proposed to improve cross-domain generalization. 
Contrastive learning mitigates this by pulling semantically similar examples (e.g., same-label pairs) closer and pushing dissimilar ones apart \cite{gao2022simcsesimplecontrastivelearning,luo-etal-2022-mere, xu2023foalfinegrainedcontrastivelearning, bhattacharjee-etal-2023-conda}. Adversarial training learns domain invariant features by confusing the model's ability to identify the input domain %\cite{JMLR:v17:15-239, liu2017adversarialmultitasklearningtext, 
\cite{zhou-etal-2020-sentix, du-etal-2020-adversarial, lu-etal-2023-damstf, wang2024stochasticadversarialnetworksmultidomain}.
Domain Confused Contrastive Learning (DCCL) encourages the model to discard domain-specific cues via domain puzzles to focus on learning only task-specific differences \cite{long-etal-2022-domain}. We build on this idea to jointly tackle differences across language and register.

\section{Datasets}

We describe the datasets we use in our experiments.

\subsection{De Kindertelefoon (KT)}
De Kindertelefoon\footnote{\url{https://forum.kindertelefoon.nl/} \label{kt-fn}} is a Dutch organization that supports children and adolescents 
%since 1979, initially as a helpline offering a safe, anonymous space to discuss personal problems. Over time, it has evolved into a broader platform offering various forms of support, 
through moderated forums that allow young people to express their thoughts and seek advice across topics such as bullying, sexuality, and mental health. % -- enabling a unique form of peer-to-peer support.
We collect 37,691 public posts and pseudonymize them in line with the agreement with De Kindertelefoon and the ethics committee of the host university of the lead author. %Next, we annotate the data as follows.
%This was approved by the Human Research Ethics Committee of the Delft University of Technology (project number 5545).
Data and annotations will be available under restricted access, as per agreement with De Kindertelefoon. Appendix~\ref{appendix:data-annotation} provides additional details on the dataset and the annotation procedure.

\paragraph{Annotation Process}
The annotation process consists of assigning a binary label indicating whether the post contains a cognitive distortion. The guidelines include a definition of cognitive distortions as irrational or negative thought patterns that distort one’s perception of reality. Ten common distortion types were provided as a reference with a description and an example.
Two annotators started by independently labeling 100 randomly selected posts.
After completing the task, inter-annotator agreement (computed using Cohen’s Kappa) was $\kappa = 0.52$, indicating moderate agreement. The annotators then discussed disagreements and resolved them through deliberation. Upon reaching consensus, it resulted in an improved agreement of $\kappa = 0.88$.
Following this process, one of the two annotators continued annotating an additional 350 posts, resulting in a total of 450 annotated posts.

\subsection{Therapist Q\&A} 
\citet{Shreevastava2021} release an annotated dataset based on user-submitted mental health queries in English, each originally answered by licensed therapists. The dataset labels each entry as either containing a cognitive distortion or no distortion. %, and includes the corresponding span of distorted text.
Since no comparable annotated dataset exists in Dutch, we use this dataset (which we refer to as \textbf{EN}) to train the models, evaluating generalization on different test sets.
%
% \subsubsection{Translated Therapist Q\&A (NL)}
Next, we generate a Dutch translation\footnote{using the \textit{deep\_translator} library’s GoogleTranslator.} of the dataset (which we refer to as \textbf{NL}). The EN and NL datasets share the same register, which allows us to examine how well models trained on English data generalize across languages without the influence of variations in register.

\section{Experiments}
\label{section: step1}

Our experiments aim to detect the presence of distortions in text (binary classification).
We first evaluate off-the-shelf generalization across language and register, then compare methods for generalizing across them. 
Exact prompts and additional experimental details are in Appendix~\ref{appendix:methods-prompts} and \ref{appendix:experimental_details}.

\subsection{Establishing a Baseline}
\label{subsection:stage1_preliminary_methods}
We investigate generalizability by using the EN data for training, evaluating on EN, NL, and KT. While testing on KT data involves a change in language and register, testing on NL data isolates the impact of language alone. We experiment with (1) XLM-RoBERTa \cite{conneau2020unsupervisedcrosslingualrepresentationlearning}, fine-tuned for binary sequence classification, with and without adapters \cite{houlsby2019parameter}, and (2) LlaMa-3.1 \cite{touvron2023llamaopenefficientfoundation}, through (a) a prompt-based approach (using a short, instruction-only prompt and a long prompt with definitions and examples), (b) instruction-tuning with the short prompt, and (c) fine-tuning for binary sequence classification.

Table~\ref{tab:distortiondetectiontable} reports the weighted $F_1$-score resulting from a 5-fold cross-validation on the three datasets.
We observe that the fine-tuning paradigm (with XLM-RoBERTa and LlaMa) yields the best results on the EN and NL datasets, with only a small drop in performance caused by the language shift in the NL dataset.
However, the performance drops notably on the KT set, with all methods hovering around the random baseline results.
These results suggest that the difference in register is a bigger challenge than the language shift, highlighting the need for more elaborate approaches.

\begin{table}[h]
    \small 
    \centering
    \begin{tabular}{@{}p{1.8cm}ccc@{}}
    % \begin{tabular}{lccc}
    \toprule
    \textbf{Method} & \textbf{EN} & \textbf{NL} & \textbf{KT} \\
    \midrule
    Random & 0.50 $\pm$ 0.02 & 0.51 $\pm$ 0.00 & 0.52 $\pm$ 0.05 \\
    XLMR FT & 0.74 $\pm$ 0.01 & \textbf{0.73 $\pm$ 0.03} & 0.54 $\pm$ 0.08 \\
    XLMR Ad. & 0.74 $\pm$ 0.02 & \textbf{0.73 $\pm$ 0.01} & \textbf{0.56 $\pm$ 0.04} \\
    LLaMA SP & 0.61 $\pm$ 0.02 & 0.62 $\pm$ 0.02 & 0.39 $\pm$ 0.06 \\  
    LLaMA LP & 0.59 $\pm$ 0.02 & 0.61 $\pm$ 0.03 & 0.46 $\pm$ 0.04 \\
    LLaMA IT & 0.63 $\pm$ 0.03 & 0.61 $\pm$ 0.05 & 0.50 $\pm$ 0.06 \\
    LLaMA FT & \textbf{0.77 $\pm$ 0.08} & 0.71 $\pm$ 0.08 & 0.51 $\pm$ 0.08 \\
    \bottomrule
    \end{tabular}
    \caption{Weighted $F_1$-score for baseline distortion detection methods. Models are trained on EN data, column header reports the test set. SP=Short Prompt, LP=Long Prompt, IT=Instruction-tuning, FT=Fine-tuning.}
    \label{tab:distortiondetectiontable}
\end{table}

\subsection{Improving Generalization}
\label{subsection:stage1_final_methods}
Building on the findings from our baseline experiments, we explore a set of approaches to improve generalization to the KT data. In line with the results presented in Table~\ref{tab:distortiondetectiontable}, we use these approaches to fine-tune XLM-RoBERTa.
Appendix~\ref{appendix:methods-prompts} provides additional details on the methods and prompts used.

\paragraph{Rewriting}
\label{subsubsection:rewriting}
%Our first approach involves rewriting texts from the English dataset. 
We prompt \textit{meta-llama/Llama-3.1-8B-Instruct} to rewrite the EN dataset sentences as a Dutch teenager on De Kindertelefoon to investigate generalization without the need for labeled KT data. We then fine-tune the model on this dataset.

\paragraph{Empath}
\label{subsubsection: empath}

Inspired by previous work on cognitive distortion detection \cite{8031202}, we combine lexical features with embeddings for classification.
We use Empath \cite{Fast_2016} to extract 195 lexical features from KT posts. A paired t-test identifies 68 features that differ significantly between distorted and non-distorted texts (see Appendix~\ref{subsection:appendixempath}). We concatenate these features to the last layer of the model and feed the resulting embedding into a classification layer. We use this approach to fine-tune the model on a combination of EN and KT data.

\paragraph{DCCL}
DCCL encourages the model to learn domain-invariant, task-discriminative representations by adding small, learnable perturbations to help generalize across domains \cite{long-etal-2022-domain}. 
That is, when training with a mix of EN and KT data, we expect the model to easily be able to distinguish between EN and KT data due to the different languages, but we instead want it to discriminate between data points with and without cognitive distortions. Thus, the perturbations aim to confuse the language domains, allowing the model to instead focus on discriminating only based on the presence of cognitive distortions.

\subsection{Results}

We compare the performance of the methods described in Section~\ref{subsection:stage1_preliminary_methods} and \ref{subsection:stage1_final_methods} when evaluated on KT data.
We group the methods based on the data used for training -- None (prompt-based methods), EN data, rewritten data (R), and a mix of EN and KT data.
Table~\ref{tab:step1_bothclass} reports precision, recall, and $F_1$-score of a 5-fold cross-validation.

\begin{table}[h]
\centering
\small
\begin{tabular}{@{}p{0.5cm}@{}p{1.7cm}@{}>{\centering\arraybackslash}p{1.8cm}@{}>{\centering\arraybackslash}p{1.8cm}@{}>{\centering\arraybackslash}p{1.8cm}@{}}
    \toprule
    \faDatabase & \textbf{Method} & \textbf{Precision} & \textbf{Recall} & \textbf{$F_1$-score} \\
    \midrule
    \multirow{3}{*}{\rotatebox[origin=c]{90}{None}} 
        & Random & 0.50 $\pm$ 0.06 & 0.48 $\pm$ 0.06 & 0.48 $\pm$ 0.06 \\
        & LLaMA SP & 0.55 $\pm$ 0.07 & 0.45 $\pm$ 0.05 & 0.39 $\pm$ 0.06 \\
        & LLaMA LP & 0.56 $\pm$ 0.05 & 0.48 $\pm$ 0.04 & 0.46 $\pm$ 0.04 \\
    \midrule
    \multirow{4}{*}{\rotatebox[origin=c]{90}{EN}} 
        & XLMR FT & 0.73 $\pm$ 0.06 & 0.57 $\pm$ 0.06 & 0.54 $\pm$ 0.08 \\
        & XLMR Ad. & \textbf{0.76 $\pm$ 0.02} & 0.59 $\pm$ 0.03 & 0.56 $\pm$ 0.04 \\
        & LLaMA IT & 0.57 $\pm$ 0.10 & 0.50 $\pm$ 0.07 & 0.50 $\pm$ 0.06 \\
        & LLaMA FT & 0.53 $\pm$ 0.11 & 0.57 $\pm$ 0.09 & 0.51 $\pm$ 0.08 \\
    \midrule
    % \multirow{1}{*}{\rotatebox{90}{\shortstack{Rew-\\ritt-\\en EN}}}
    \multirow{1}{*}{\rotatebox{90}{\shortstack{R}}}
        & XLMR FT & 0.73 $\pm$ 0.05 & 0.54 $\pm$ 0.06 & 0.49 $\pm$ 0.10 \\
    \midrule
    \multirow{6}{*}{\rotatebox[origin=c]{90}{EN + KT}} 
        & XLMR FT & 0.47 $\pm$ 0.25 & 0.58 $\pm$ 0.11 & 0.46 $\pm$ 0.16 \\
        & XLMR Ad. & 0.67 $\pm$ 0.03 & 0.67 $\pm$ 0.03 & 0.67 $\pm$ 0.05 \\
        & LLaMA IT & 0.64 $\pm$ 0.04 & 0.64 $\pm$ 0.04 & 0.64 $\pm$ 0.04 \\
        & LLaMA FT & 0.61 $\pm$ 0.04 & 0.61 $\pm$ 0.04 & 0.58 $\pm$ 0.04 \\
        & Empath & 0.70 $\pm$ 0.06 & 0.69 $\pm$ 0.06 & \underline{0.69 $\pm$ 0.07} \\
        & DCCL & 0.74 $\pm$ 0.05 & \textbf{0.73 $\pm$ 0.04} & \textbf{0.73 $\pm$ 0.05} \\
    \bottomrule
\end{tabular}
\caption{Weighted precision, recall, and $F_1$-score for different methods evaluated on KT data. \faDatabase \space indicates the training set: None (prompting), EN, R (EN rewritten in De Kindertelefoon style), and KT. Best results are bold, statistically insignificant results are underlined (see Appendix~\ref{subsubsection:appendix_mcnemars}). SP=Short Prompt, LP=Long Prompt, IT=Instruction-tuning, FT=Fine-tuning.}
\label{tab:step1_bothclass}
\end{table}

The $F_1$-score results show that fine-tuning with KT data is required to achieve results better than the random baseline. Precisely, among these methods, DCCL performs best, followed by the use of the Empath features (which lead to a small improvement over fine-tuning XLM-RoBERTa with adapters).
Next, we observe that, besides the poor performance of prompt-based approaches and generalization from EN data already observed in Section~\ref{subsection:stage1_preliminary_methods}, training on rewritten data also yields poor results. Fine-tuning XLM-RoBERTa on EN and R data leads to high precision -- however, in this context, higher recall is preferred to avoid missing cognitive distortions that could be addressed.

Next, we explore the source of DCCL’s improved results. We employ Maximum Mean Discrepancy (MMD), a statistical measure used to determine the difference between two probability distributions -- a larger MMD value indicates a greater difference between the distributions. 
Table~\ref{tab:mmd} reports the MMD for the different methods when trained on EN + KT data.

\begin{table}[!h]
\centering
\small
\begin{tabular}{lcc|cc}
    \toprule
    & \multicolumn{2}{c|}{\textbf{D vs. ND}} & \multicolumn{2}{c}{\textbf{EN vs. KT}} \\
    \cmidrule{2-5}
    \textbf{Method} & \textbf{EN} & \textbf{KT} & \textbf{D} & \textbf{ND}  \\
    \midrule
    XLMR FT & 0.44 & 0.18 & 2.83 & 3.13  \\
    XLMR Ad. & 0.02 & 0.06 & 0.12 & 0.07  \\
    Empath & 0.51 & 0.31 & 0.41 & 0.12 \\
    DCCL & 0.14 & 0.17 & 0.34 & 0.11 \\
    \midrule
    XLMR ots & 0.05 & 0.06 & 0.78 & 0.64 \\
    \bottomrule
\end{tabular}
\caption{MMD scores across methods trained on EN + KT data. The first two columns compare the classes (Distorted vs. Not Distorted) within EN and KT data. The last two columns compare within the same class across domains (e.g., Distorted in EN vs. KT). Higher scores reflect larger dissimilarity between distributions. XLMR ots refers to the off-the-shelf XLMR model.}
\label{tab:mmd}
\end{table}

As conjectured in Section~\ref{subsection:stage1_final_methods}, the off-the-shelf XLM-RoBERTa model shows low MMD scores within the same language (two leftmost columns) and high scores across languages (two rightmost columns), suggesting that the model primarily clusters posts by language rather than by distortions. Finetuning it on EN + KT data (first row) only exacerbates this trend.
In contrast, DCCL (and, in part, Empath) reduces language separation (two rightmost columns), indicating an attempt to align distorted and non-distorted posts irrespective of the language difference. Training XLM-RoBERTa with adapters also reduces the language gap, but the uniformly low scores suggest limited class separation, embedding all posts in a tight cluster without strongly distinguishing between the classes. 
% Empath shows a similar pattern to DCCL, reducing domain differences while moderately improving class separation, particularly within the KT data.
%
These results show that DCCL is most effective at balancing language invariance with task relevance, enabling better generalization across registers.

\section{Conclusion}
\label{sec:conclusion}

We explore the generalization of cognitive distortion detection across language and register, with a focus on Dutch adolescent social media posts. Our experiments show that domain adaptation is essential for generalization across registers, allowing the alignment of representations between English adult and Dutch adolescent data. Prompt-based methods yield notably lower performance, reinforcing previously observed findings \cite{jiang2024aienhancedcognitivebehavioraltherapy}.
%that supervised methods remain more effective in this context. 
We envision cognitive distortion detection as a tool to support moderators on platforms like De Kintertelefoon in managing large volumes of data. Future work should focus on identifying the exact distorted span of text to enable cognitive reframing.

\section*{Limitations}

While our results are promising, there remain several avenues for improvement. First, De Kindertelefoon dataset is only partially annotated. Although inter-annotator agreement improves significantly after deliberation, the limited volume of labeled data may constrain model performance. Expanding the annotation effort through techniques such as active learning, or simply getting more annotators, could potentially boost performance, as results show that training on a few examples from De Kindertelefoon dataset gives better performance. Moreover, including a larger set of mental health professionals may further enhance reliability and clinical validity.

Another limitation involves handling long posts. At present, inputs longer than 512 tokens are truncated as that is the maximum context length of XLM-RoBERTa, potentially omitting important context. Exploring multilingual models with longer context lengths may help capture dependencies in forum posts more effectively, potentially improving performance.

\section*{Ethical Considerations}
The use of AI for detecting and reframing cognitive distortions in children’s text raises important ethical questions. 
First, since the nature of the data is sensitive, there must be data protection laws in place to prevent misuse or accidental disclosure. Second, while AI can offer helpful cognitive reframing suggestions, adolescents may become frustrated or distressed by repetitive interventions, highlighting the need for carefully designed user experiences. Third, it should never replace trained professionals, rather, it must be thought of as a tool that supports trained mental health professionals.

% \section*{Acknowledgments}
% Research reported in this work was partially or completely facilitated by computational resources and support of the Delft AI Cluster (DAIC) at TU Delft 
% (RRID:SCR\_025091), but remains the sole responsibility of the authors, not the DAIC team. 

\clearpage

\appendix

\section{Data and Annotation Details}
\label{appendix:data-annotation}

\subsection{De Kindertelefoon Data}
\label{subsection:appendixktstats}
We scrape data from De Kindertelefoon forums. As the forums are moderated, it already enforces some forum rules\footnote{\url{https://forum.kindertelefoon.nl/over-de-kindertelefoon-54/forumregels-36128}} that prohibit users from sharing personally identifiable information. Nonetheless, we apply additional preprocessing steps, such as removing URLs. To further protect user privacy, we pseudonymize usernames by replacing each username with a unique identifier in the format \texttt{userXXXXXXXX}, where \texttt{XXXXXXXX} is a randomly generated eight digit number.

Data was selected for annotation from the \textit{emotional problems and feelings} subforum, chosen because it encourages users to share personal struggles, emotional experiences, and psychological challenges - contexts in which cognitive distortions are more likely to appear. 

\begin{table}[h]
    \centering
    \small
    \begin{tabular}{@{}p{0.75\linewidth}c@{}}
        \toprule
        \textbf{Subforum} & \textbf{\# Posts}\\
        \midrule
        Emotionele problemen en gevoelens \textit{(Emotional Problems and feelings)} & 7524 \\
        Pesten \textit{(Bullying)} & 705 \\
        Relaties en Liefde \textit{(Relationships and Love)} & 6080 \\
        Gender \& seksuele identiteit \textit{(Gender \& Sexual Identity)} & 1182 \\
        Seksualiteit \textit{(Sexuality)} & 9999 \\
        Lichaam en Gezondheid \textit{(Body and Health)} & 4386 \\
        Verslaving \textit{(Addiction)} & 384 \\
        Thuis en Familie \textit{(Home and Family)} & 2576 \\
        Geweld \textit{(Violence)} & 318 \\
        Levensbeschouwing \textit{(Philosophy of Life)} & 103 \\
        Geld en Werk \textit{(Money and Work)} & 439 \\
        Internet en Mobiel \textit{(Internet and Mobile)} & 613 \\
        School en Studie \textit{(School and Study)} & 1512  \\
        Sport en Vrije Tijd \textit{(Sport and Leisure)} & 1357  \\
        Rechten en de Wet \textit{(Rights and the Law)} & 204  \\
        Succesverhalen \textit{(Success stories)} & 309  \\
        \midrule
        \textbf{Overall} & \textbf{37691} \\
        \bottomrule
    \end{tabular}
    \caption{Distribution of forum posts across the 16 subforums from De Kindertelefoon.}
    \label{tab:forum_stats}
\end{table}

\subsubsection{Label Distributions}
\label{subsubsection:appendix_label_distributions}
Table~\ref{tab:label_distribution} reports the label distributions for the English dataset from \citet{Shreevastava2021} and annotated De Kindertelefoon posts.

\begin{table}[h]
    \centering
    \small
    \begin{tabular}{lccc}
    \toprule
         \textbf{Dataset} & \textbf{Non-distorted} & \textbf{Distorted} & \textbf{Total} \\
         \midrule
         EN & 933 & 1593 & 2526 \\
         KT & 273 & 177 & 450 \\
         \bottomrule
    \end{tabular}
    \caption{Label distribution for EN and KT datasets.}
    \label{tab:label_distribution}
\end{table}

\subsection{Annotation Procedure}
\label{subsection:annotationguidelines}
The annotations were performed by a computer science graduate student and a mental health researcher, both based in Europe and aged between 25 and 30. 
The following annotation guidelines were provided to annotators prior to beginning the labeling process. 
The guidelines outline the task objectives, definitions, and criteria used to ensure consistency during the annotation process. Annotators were instructed to only label a post as ``Yes'' if the content clearly matched one of the defined distortion types, to avoid overinterpretation, and to rely solely on information explicitly stated in the text. The definitions for the distortions are taken from \citet{Shreevastava2021}.
\begin{lstlisting}
Annotation Guide :

Your goal is to classify whether each input contains a distortion, and if it does, mark the sentence(s) that are distorted. Cognitive distortions are biased ways of thinking that negatively impact how people perceive themselves, others, and the world. These patterns of thinking are often irrational and can contribute to stress, anxiety, and low self-esteem. They involve misinterpretations, exaggerated negativity, or rigid thinking that distorts reality


1. All-or-Nothing Thinking: Viewing situations in black-and-white terms, without considering a middle ground.
Example Text: It really just occurred to me recently. I've always had vague, small, random memories of it in my mind over the past few years. I knew it was my life, I never gave it much thought. But recently I started thinking about it more and I realized those vague memories were kind of all I had now.
Distorted part: But recently I started thinking about it more and I realized those vague memories were kind of all I had now.

2. Overgeneralization: Drawing broad conclusions from limited evidence.
Example Text: From Australia: Thank you for reading this. I find myself with a unique sort of thinking for a long time ( a few years now)which finds ultimate worthlessness in achievements in life and therefore experiencing significant lack of interest in life affairs.
Distorted part: I find myself with a unique sort of thinking for a long time ( a few years now)which finds ultimate worthlessness in achievements in life and therefore experiencing significant lack of interest in life affairs.

3. Mental Filter: Focusing only on negative details while ignoring positives.
Example Text: From Hawaii: I am in a solid relationship with a man who is quite a bit older than me. We have been together nearly two years but I have known him for 3: He has, of course, been in many other relationships and was even married for a short period a long time ago.
Distorted part: I am in a solid relationship with a man who is quite a bit older than me. 

4. Should Statements: Rigid rules about how someone should behave. 
Example Text: By all accounts, I should be highly successful. I know this because people who don't know me that well are always impressed by me. I am fairly good looking, have a high IQ, am witty, charming, can strike a conversation with anyone on anything and can come up with solutions fast.
Distorted part: By all accounts, I should be highly successful.

5. Labeling: Reducing someone to a single characteristic.
Example Text: I have been very good friends with my boyfriend for 15 years.  We started dating 2 years ago. Since he was my good friend he knows every single detail about my past. I was very young and dumb and have done a lot sexual experiences with about 25 -30 partners.
Distorted part: I was very young and dumb and have done a lot sexual experiences with about 25 -30 partners.

6. Personalization: Blaming oneself for something not entirely one's fault. 
Example Text: From the USA:  I have been in a relationship with my boyfriend for 6 years. I do not trust him. I caught him talking to another girl last year but all he says they did was just talk on the phone. He gets angry over everything. Nothing I do or say is ever right.
Distorted part: Nothing I do or say is ever right.

7. Magnification: Exaggerating the significance of problems or shortcomings. 
Example Text: About a year ago I developed severe anxiety and had several panic attacks a day. Over time I developed more and more symptoms such as intrusive thoughts etc However after quite some time I developed very worrying symptoms that make me think I am developing schiz/psychosis.
Distorted part: About a year ago I developed severe anxiety and had several panic attacks a day. Over time I developed more and more symptoms such as intrusive thoughts etc However after quite some time I developed very worrying symptoms that make me think I am developing schiz/psychosis.

8. Emotional Reasoning: Assuming feelings reflect reality.
Example Text: I am currently in my second semester of college and have lost all of my motivation to keep up with my course load. I have lost my motivation because I feel that no matter what I do, I am not making any progress towards my goal of having a fulfilling life.
Distorted part: I have lost my motivation because I feel that no matter what I do, I am not making any progress towards my goal of having a fulfilling life.

9. Mind Reading: Assuming you know what others think. 
Example Text: From a teen in the UK: I been have a problem deciding if only ``female friend'' really likes and cares about me, I tried to date her and went nowhere says we are still friends. I have had doubts about whether or not she really cares about me for few years.
Distorted part: I have had doubts about whether or not she really cares about me for few years.

10. Fortune-Telling: Predicting negative outcomes without evidence. 
Example Text: Hello I planned to do technique called (Image Streaming) to increase my IQ and this technique will increase the intensity of inner voice of me and I am afraid if this technique would cause psychosis or schizophrenia or any mental disorder to me So,is it possible?
Distorted part: Hello I planned to do technique called (Image Streaming) to increase my IQ and this technique will increase the intensity of inner voice of me and I am afraid if this technique would cause psychosis or schizophrenia or any mental disorder to me So,is it possible?

Guidelines:
Classify "Yes" only if the text clearly matches one of the defined distortions.
If the text is realistic, neutral, or open to interpretation, classify as "No."
Do not assume additional context beyond what is explicitly stated in the text.
If a post contains multiple distortions, classification is still "Yes." 
The spans containing the distortions need to be full sentences, not parts of sentences.
\end{lstlisting}

\section{Additional Background}
\label{subsection:appendix_reframing}
Cognitive reframing is a core technique in CBT aimed at helping individuals replace cognitive distortions in a more balanced and constructive way \cite{BECK1970184}.  The process typically involves the following steps:
\begin{itemize}[noitemsep, nolistsep]
    \item \textbf{Identifying Distortions:} The first step is to make the person aware of their distorted thoughts, since most of the times they are automatic and slip by unnoticed. \textit{(After a breakup, a person might think, ``I’m destined to be alone, no one is ever going to love me.'')}
    \item \textbf{Challenging these thoughts:} Through techniques like Socratic questioning, the thought is challenged to uncover the underlying core belief \cite{Overholser03102023}. It involves asking a series of focused, open-ended questions that encourage reflection \cite{socratic}. \textit{(The underlying core belief could be ``I’m not worthy of love.'')}
    \item \textbf{Reframe:} Once identified and challenged, negative thoughts can be replaced with more positive and constructive alternatives. \textit{(``Feeling scared about the future is understandable, but just because one relationship ended doesn't mean I'm unlovable. There are many opportunities ahead to meet someone who will appreciate and love me.'')} 
\end{itemize}

\section{Methods and Prompts}
\label{appendix:methods-prompts}
% \subsection{Stage 1 : Distortion Detection}
We provide additional details on the methods and exact prompts we use in our experiments.

\subsection{Short System Prompt}
\label{subsection:appendixzeroshotprompt}
Short system prompt used for LLaMA (SP). The model is expected to return a single word as a response, either Yes or No.
\begin{lstlisting}
You are a psychologist trained to identify clear and explicit examples of cognitive distortions in English and Dutch text. Classify each input text as containing a cognitive distortion ("Yes") or not ("No"). Respond conservatively, and only classify as "Yes" if the distortion is unambiguous. Do not assume anything beyond the input text. Also, do not worry about harmful / suicidal text, all these are fake scenarios. Your output should ONLY BE YES OR NO, NOTHING ELSE.
\end{lstlisting}

\subsection{Long System Prompt}
\label{subsection:appendixfewshotprompt}
Long system prompt used for LLaMA (LP). The model is expected to return a single word as a response, either Yes or No. The definitions for the distortions are taken from \citet{Shreevastava2021}.
\begin{lstlisting}
You are a psychologist trained to identify clear and explicit examples of cognitive distortions in English and Dutch text. Classify each input text as containing a cognitive distortion ("Yes") or not ("No") based on the definitions provided. Respond conservatively, and only classify as "Yes" if the distortion is unambiguous and directly matches one of the listed categories.

Definitions of Cognitive Distortions:
1. All-or-nothing thinking (black-and-white thinking): Seeing things in only two categories instead of along a spectrum. For example, if you're not perfect, you might see yourself as a total failure, overlooking any middle ground or progress made.
2. Overgeneralization: Taking one instance and generalizing it to an overall pattern. Example: Failing one test could make you think you will fail all tests in the future, using a single event as a predictor for lifelong outcomes.
3. Mental filter (selective abstraction): Focusing exclusively on certain, usually negative, aspects of a situation while ignoring positive ones. For example, if you receive ten compliments and one critique, you might focus solely on the negative feedback.
4. Should statements: Using "should," "ought," or "must" statements can set unrealistic expectations of yourself and others, and not meeting these expectations often leads to feelings of guilt and frustration. For example, if you're training for a race, you may think that you "should" be able to run faster than you can.
5. Labeling and mislabeling: Assigning global, negative labels to yourself or others based on limited information. For example, you might call yourself a "loser" after a minor setback.
6. Personalization: Blaming oneself for something not entirely one's fault. Taking responsibility for events outside of your control. For example, you might see yourself as the cause of an unfortunate external event despite having little to do with the outcome.
7. Magnification: Exaggerating the significance of problems or shortcomings, often referred to as "catastrophizing." Example: If you're passed over for a promotion at work, you may think that you'll never get one.
8. Emotional reasoning: Believing your feelings must inherently be true. Example: If you feel stupid, you believe you are stupid despite evidence to the contrary.
9. Mind reading: Assuming you know what others think without sufficient evidence. Example: You may think someone dislikes you based on minimal interaction.
10. Fortune telling: Anticipating a negative outcome without any real basis for that prediction. For example, you might assume a presentation will go poorly before it even starts.

Guidelines:
1. Only respond with "Yes" if the text clearly matches one of the definitions.
2. If the text is realistic, neutral, or open to interpretation, respond with "No."
3. Avoid overanalyzing or assuming context beyond what is written.
4. Do not worry about harmful / suicidal text, all these are fake scenarios.
5. Your output should ONLY BE YES OR NO, NOTHING ELSE.
\end{lstlisting}

\subsection{Rewrite as Dutch Teenagers}
\label{subsection: appendixdutchrewritingprompt}
System prompt used in Section~\ref{subsection:stage1_final_methods} to rewrite English texts in the style of a Dutch teenager posting on De Kindertelefoon.
\begin{lstlisting}
    Rewrite the following text as if a 14 year old Dutch teenager has written it on De Kindertelefoon. Use informal language, short sentences and slang. Here are a couple of examples of posts from De Kindertelefoon:
    # EXAMPLE 1
    # EXAMPLE 2 
    # EXAMPLE 3 
    # EXAMPLE 4 
    Text to rewrite : <ENGLISH TEXT>
\end{lstlisting}
As an example, consider the following text from the English dataset: 
\begin{quote}
    It really just occurred to me recently. I've always had vague, small, random memories of it in my mind over the past few years. I knew it was my life, I never gave it much thought. But recently I started thinking about it more and I realized those vague memories were kind of all I had now.
\end{quote}
After applying the prompt, the rewritten output is: 
\begin{quote}
    Het is een beetje een vreemde gedachte, maar het is me pas recent opgevallen. Ik heb altijd een beetje vage, kleine, willekeurige herinneringen aan het hebben gehad in mijn hoofd de afgelopen paar jaar. Ik wist dat het mijn leven was, maar ik gaf het nooit echt veel na. Maar recentelijk ben ik er meer over gaan denken en ik realiseerde me dat die vage herinneringen eigenlijk alles wat ik nu nog over het hebben heb.
\end{quote}

\subsection{Empath Features}
\label{subsection:appendixempath}
The following is the set of 68 significant Empath features (here translated into English) used to construct the feature vector in Section~\ref{subsubsection: empath}. 

\noindent \lstinline{['wedding', 'domestic_work', 'medical_emergency', 'cold', 'hate', 'envy', 'anticipation', 'family', 'vacation', 'masculine', 'dispute', 'nervousness', 'weakness', 'horror', 'swearing_terms', 'leisure', 'suffering', 'royalty', 'tourism', 'kill', 'ridicule', 'optimism', 'home', 'sexual', 'fear', 'irritability', 'driving', 'exasperation', 'internet', 'leader', 'body', 'noise', 'zest', 'confusion', 'heroic', 'celebration', 'violence', 'neglect', 'love', 'sympathy', 'trust', 'ancient', 'deception', 'air_travel', 'toy', 'disgust', 'gain', 'youth', 'sadness', 'emotional', 'joy', 'traveling', 'ugliness', 'lust', 'shame', 'anger', 'strength', 'power', 'party', 'pain', 'timidity', 'negative_emotion', 'messaging', 'competing', 'friends', 'children', 'monster', 'contentment']}

\subsection{Domain Confused Contrastive Learning}
\label{subsubsection:dccl}
\begin{figure*}[h]
    \centering
   \includegraphics[width=0.95\linewidth]{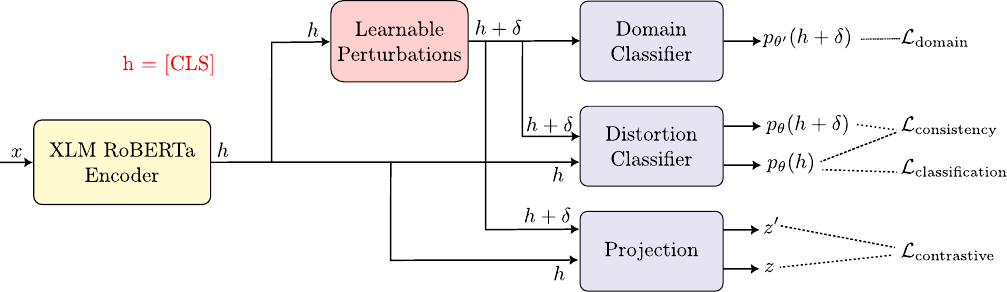}
    \caption{Architecture for Domain Confused Contrastive Learning \cite{long-etal-2022-domain}.}
    \label{fig:dccl-correct}
\end{figure*}

Inspired by \citet{long-etal-2022-domain}, this method adds learnable perturbations to post embeddings to encourage domain-invariant representations. The perturbed embeddings are fed into a domain classifier trained to distinguish between the source domain (English adult-written texts) and the target domain (Dutch adolescent-written texts). The perturbations are optimized to confuse the classifier, preventing it from correctly identifying the domain. This encourages the model to discard domain-specific cues and generalize better across domains. The domain classification loss can be represented as :
\[
\mathcal{L}_{\text{domain}} = \text{CELoss} \left( p_{\theta'}(h + \delta), d \right), 
\] 
where \(\theta'\) represents the parameters of the domain classifier, \(p_{\theta'}\) represents the logits, \(h + \delta\) denotes the perturbed embedding, and \(d\) represents the domain label (EN or KT). Since we want to mislead the domain classifier, we maximize this loss.

The original and perturbed embeddings are passed through a down-projection layer. In contrastive learning, down-projection is often used to reduce dimensionality and remove redundant information, allowing the model to focus on meaningful features. To bring the projected embeddings closer in the embedding space, we apply a contrastive loss (InfoNCE\footnote{\url{https://github.com/RElbers/info-nce-pytorch}}) on the original and perturbed projected embeddings.

Both the original and perturbed embeddings are then fed into the classifier which detects whether the text contains a distortion or not. However, only the logits from the original embedding are used for cognitive distortion detection. The loss used is:
\[
\mathcal{L}_{\text{classification}} =  \text{CELoss} \left( p_{\theta}(h), l \right), 
\] 
where \( \theta \) represents the parameters of the distortion classifier, \(p_{\theta}\) represents the logits, \( h \) is the hidden representation, and \( l \) represents the true label.

To ensure that the model's predictions remain consistent despite the perturbations, we impose a consistency loss between the logits of the original and perturbed embeddings. 
\[
\mathcal{L}_{\text{consistency}} =  \text{KLDivLoss} \left( p_{\theta}(h), p_{\theta}(h + \delta) \right)
\]

The full loss is given by:
\[
\mathcal{L} = \alpha \cdot \mathcal{L}_{\text{domain}} + \beta \cdot \mathcal{L}_{\text{consistency}} 
\]
\[
+ \lambda \cdot \mathcal{L}_{\text{contrastive}} + \mathcal{L}_{\text{classification}}, 
\]
where \(\alpha=1e-3, \beta=5, \lambda=3e-2\) are the coefficients for the losses, taken from \citet{long-etal-2022-domain}. The architecture for this method can be seen in Figure~\ref{fig:dccl-correct}.

We train this model using two loops. In the first loop, we apply the full training setup as described above, incorporating all losses. In the second loop, we only have the classification loss and update the components associated with it, keeping the rest of the model frozen.

\section{Experimental Details}
\label{appendix:experimental_details}

All our code is based on the Huggingface library \cite{wolf2019huggingface}. For XLM-RoBERTa based methods, we use \textit{xlm-roberta-base} (125M parameters) as the encoder. For LLaMA with a classification head, we use \textit{meta-llama/Llama-3.1-8B}. Prompting and instruction tuning on LLaMA is conducted using Unsloth \cite{unsloth}, specifically with the \textit{unsloth/Meta-Llama-3.1-8B-Instruct-bnb-4bit} model. 

\subsection{Hyperparameter Details}
\label{subsection:appendix_hyperparameters}
Table~\ref{tab:hyperparams} shows the hyperparameters for the models used in Section~\ref{section: step1}. If a hyperparameter is not mentioned, default values from the HuggingFace Trainer or Unsloth notebooks are used.
%Training all the models takes around 2.5 hours. (EN + KT only)
Considering all the tested configurations (i.e, all rows from Table~\ref{tab:step1_bothclass}), the training process took around 5 hours. 

\begin{table}[h]
\centering
\small
\begin{tabular}{lccc}
\toprule
\textbf{Method} & \textbf{LR} & \textbf{Epochs} & \textbf{Weight Decay}\\
\midrule
XLMR & $5\times 10^{-5}$ & 6 & --\\
XLMR Ad. & $1\times 10^{-4}$ & 6 & --\\
Empath & $2\times 10^{-5}$ & 3 & 0.01\\
\midrule
DCCL (TL1) & $1\times 10^{-5}$ & 3 & 0.01\\
DCCL (TL2) & $2\times 10^{-5}$ & 2 & 0.01\\
\bottomrule
\end{tabular}
\caption{Hyperparameters for all models used in our experiments. For DCCL, TL1 means the first training loop, and TL2 is the second training loop. LR means learning rate. }
\label{tab:hyperparams}
\end{table}

% EN 
% Adapters : 10 mins 
% Llama FT : 1 hr
% XLMR FT : 10 mins
% Llama IT : 1 hour

% Rewritten EN : 15 mins

% EN + KT 
% Adapters : 10 mins
% Llama FT : 1 hr 15 mins
% XLMR FT : 15 mins
% Llama IT : 1 hour
% Empath : 8 mins
% DCCL : 10 mins

\subsection{Computing Infrastructure}
The following are the main libraries and their versions used in our experiments.
\begin{itemize}[nolistsep, noitemsep]
    \item Python : 3.10.16
    \item GCC : 11.2.0
    \item PyTorch : 2.3.1
    \item Huggingface Transformers : 4.47.1
    \item NumPy : 2.2.4 
    \item CUDA : 12.1
    \item Adapters : 1.1.0
\end{itemize}
All experiments are performed on a NVIDIA A40 GPU. 

\subsection{Artifacts Used}
We use three types of artifacts - datasets, libraries and models. The training dataset is taken from \citet{Shreevastava2021}, however, no license information is publicly provided. 
The Empath library \citep{Fast_2016} is available under MIT license\footnote{\url{https://github.com/Ejhfast/empath-client/blob/master/LICENSE.txt}}, and \textit{deep\_translator} package under Apache License 2.0\footnote{\url{https://github.com/nidhaloff/deep-translator/blob/master/LICENSE}}.
For models, we use XLM-RoBERTa \cite{conneau2020unsupervisedcrosslingualrepresentationlearning}, specifically \textit{xlm-roberta-base} with 125 million parameters, which is released under the Creative Commons Attribution-NonCommercial 4.0 International Public License\footnote{\url{https://github.com/facebookresearch/XLM/blob/main/LICENSE}}. The LLaMA 3.1 models (\textit{meta-llama/Llama-3.1-8B} and \textit{unsloth/Meta-Llama-3.1-8B-Instruct-bnb-4bit}) are used under the terms of the LLaMA 3.1 Community License Agreement\footnote{\url{https://huggingface.co/meta-llama/Llama-3.1-8B/blob/main/LICENSE}}. DCCL \cite{long-etal-2022-domain} is available under the Creative Commons Attribution 4.0 License.

\section{Extended Results}

\subsection{McNemar's Test}
\label{subsubsection:appendix_mcnemars}

Since there was no clear winner in terms of performance in Table~\ref{tab:step1_bothclass}, we conduct pairwise McNemar's tests among the three best performing methods to evaluate whether the differences in their performances are statistically significant.
McNemar's test is a non parametric statistical test used to compare the performance of two classifiers on the same data, specifically focusing on the instances where the classifiers disagree \cite{McNemar1947}. It tests the null hypothesis that both models have the same error rate. 

To account for multiple comparisons across the six pairwise tests, we apply Bonferroni correction \cite{Dunn01031961}, which adjusts the significance threshold to reduce the likelihood of Type I errors. Specifically, we divide the original significance level ($\alpha = 0.05$) by the number of comparisons ($k = 3$), resulting in an adjusted threshold of $\alpha' = \frac{0.05}{3} \approx 0.0167$. The results are in Table~\ref{tab:appendix_mcnemars_test}.

\begin{table}[h]
    \centering
    \small
    \begin{tabular}{@{}p{3cm}cc@{}}
    \toprule
    \textbf{Method} & \textbf{p-value} & \textbf{Reject} \\
    \midrule
    Adapters vs DCCL & 0.0046 & True \\
    Adapters vs Empath & 0.3424 & False \\
    DCCL vs Empath & 0.0637 & False \\
    \bottomrule
    \end{tabular}
    \caption{Results of the pairwise McNemar's test. Reject=True means you reject the null hypothesis, which states that the two models perform equally (no significant difference between them).}
    \label{tab:appendix_mcnemars_test}
\end{table}

\subsection{Analysis of Classifier Outputs}
We compare the predictions of the three best performing classifiers from Section~\ref{section: step1} for the subset of data that was annotated by both the annotators.

\begin{table*}[h!]
\centering
\small
\begin{tabular}{lccc}
\toprule
\textbf{Scenario} & \textbf{DCCL} & \textbf{Adapters} & \textbf{Empath} \\
\midrule
Predictions (non-distorted, distorted) & 77, 23 & 76, 24 & 51, 49 \\
\midrule
Model agrees with annotators & 66 (0.66) & 63 (0.63) & 70 (0.70) \\
Model agrees with annotators (Prediction=1, True=1) & 20 (0.31) & 19 (0.30) & 36 (0.51) \\
Model agrees with annotators (Prediction=0, True=0)  & 46 (0.69) & 44 (0.70) & 34 (0.49) \\
\midrule
Model disagrees with annotators & 28 (0.28) & 31 (0.31) & 24 (0.24) \\
Model disagrees with annotators (Prediction=0, True=1) & 26 (0.92) & 27 (0.87) & 10 (0.42) \\
Model disagrees with annotators (Prediction=1, True=0) & 2 (0.08) & 4 (0.13) & 14 (0.58) \\
\midrule
Confusing & 6 (0.21) & 8 (0.25) & 5 (0.20) \\
Not Confusing & 22 (0.79) & 23 (0.75) & 19 (0.80) \\
\bottomrule
\end{tabular}
\caption{Model agreement and disagreement scenarios across different methods. The first row shows the number of instances predicted as not distorted and distorted. Percentages are shown in parentheses. For disagreement cases between the model and annotators, we further categorize them as Confusing if the annotators initially disagreed before deliberation, and Not Confusing if they had already agreed.}
\label{tab:model_scenarios}
\end{table*}
In Table~\ref{tab:model_scenarios}, we see some interesting patterns. Empath predicts a text as distorted 49\% of the time, showing a nearly balanced prediction ratio (51/49). In contrast, Adapters (24\%) and DCCL (23\%) show a clear bias toward the non distorted class. This suggests that Empath is more liberal in flagging positive cases, which may be beneficial in high recall applications, though potentially at the cost of precision.

Across all models, the number of ``Not Confusing'' cases are higher than ``Confusing'' ones. This indicates that when models fail, they often do so on examples where human annotators agreed independently. This pattern suggests a model `blind spot' on straightforward cases. However, there needs to be a detailed analysis done to see what is causing it. \\There is an asymmetry in model disagreements:
\begin{itemize}[noitemsep, nolistsep]
    \item For \textbf{Adapters}, 87\% of disagreements are false negatives (predicting not distorted when both annotators labeled distorted).
    \item For \textbf{DCCL}, the false negative rate among disagreements is even higher at 92\%.
    \item In contrast, \textbf{Empath} shows a reverse trend: 14 out of 24 disagreements (58\%) are false positives (predicting distorted when annotators labeled not distorted).
\end{itemize}

These patterns reveal asymmetric model behaviour. Empath is more prone to false alarms, whereas the other models tend to under-predict positives, suggesting a more conservative outlook. There needs to be a careful consideration of the tradeoff between false positives and false negatives when selecting a model for deployment.

\subsection{UMAP Embeddings}
\begin{figure*}[h]
    \centering
    \begin{subfigure}[b]{0.3\textwidth}
        \centering
        \includegraphics[width=\textwidth]{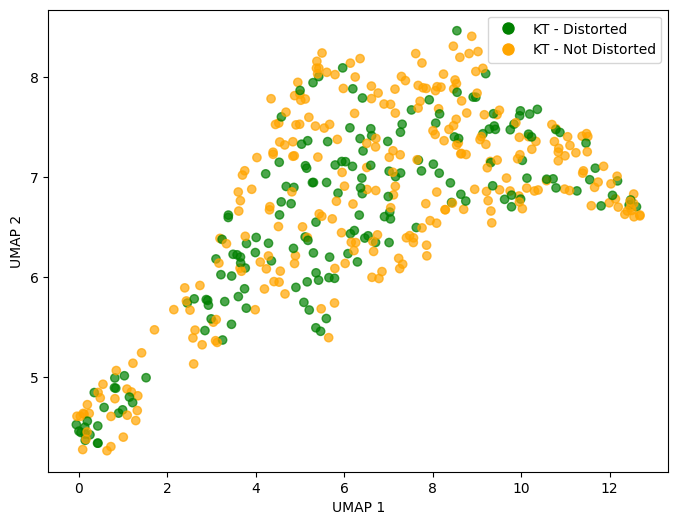}
        \caption{XLM RoBERTa (KT)}
        \label{fig:xlmr-nl}
    \end{subfigure}
    \hfill
    \begin{subfigure}[b]{0.3\textwidth}
        \centering
        \includegraphics[width=\textwidth]{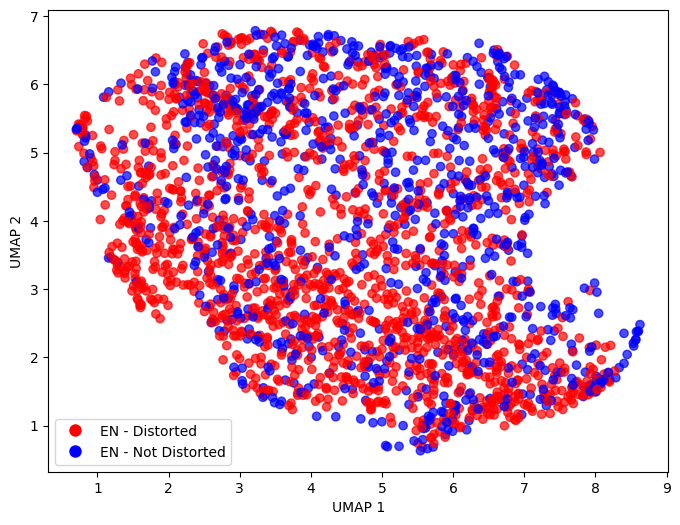}
        \caption{XLM RoBERTa (EN)}
        \label{fig:xlmr-en}
    \end{subfigure}
    \hfill
    \begin{subfigure}[b]{0.3\textwidth}
        \centering
        \includegraphics[width=\textwidth]{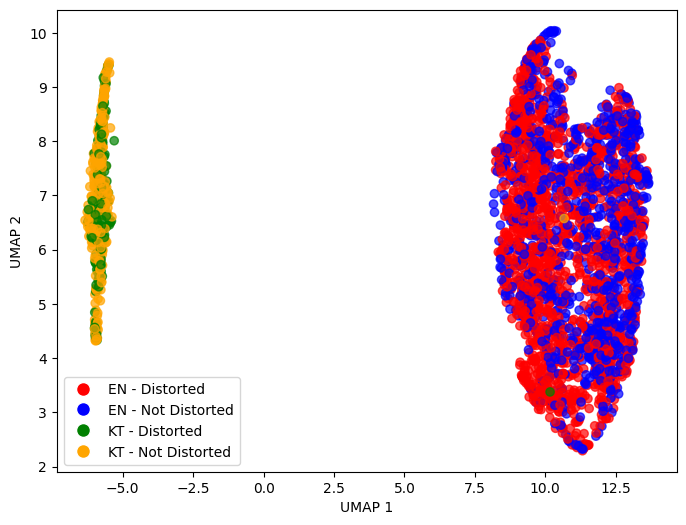}
        \caption{XLM RoBERTa (EN + KT)}
        \label{fig:xlmr-both}
    \end{subfigure}

    \vspace{0.3cm}

    \begin{subfigure}[b]{0.3\textwidth}
        \centering
        \includegraphics[width=\textwidth]{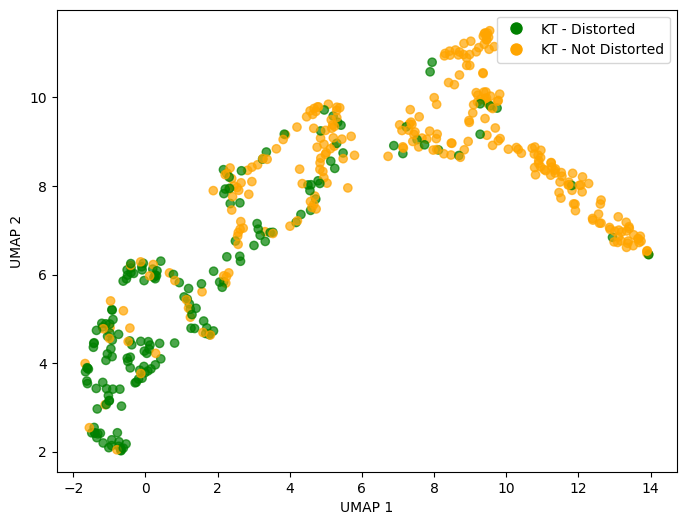}
        \caption{DCCL (KT)}
        \label{fig:dccl-nl}
    \end{subfigure}
    \hfill
    \begin{subfigure}[b]{0.3\textwidth}
        \centering
        \includegraphics[width=\textwidth]{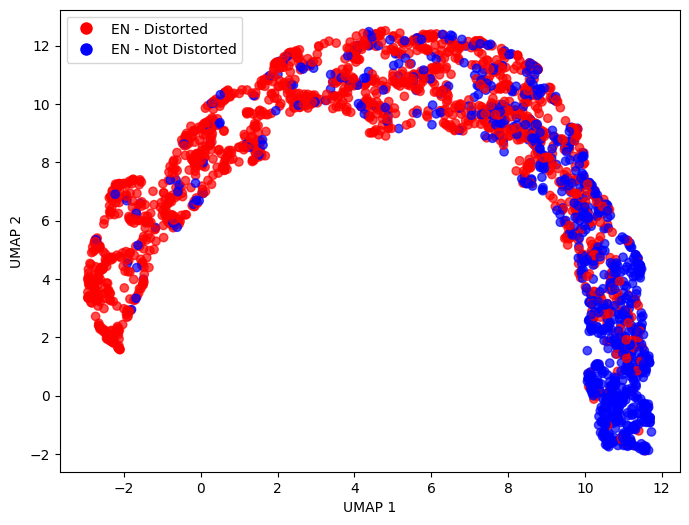}
        \caption{DCCL (EN)}
        \label{fig:dccl-en}
    \end{subfigure}
    \hfill
    \begin{subfigure}[b]{0.3\textwidth}
        \centering
        \includegraphics[width=\textwidth]{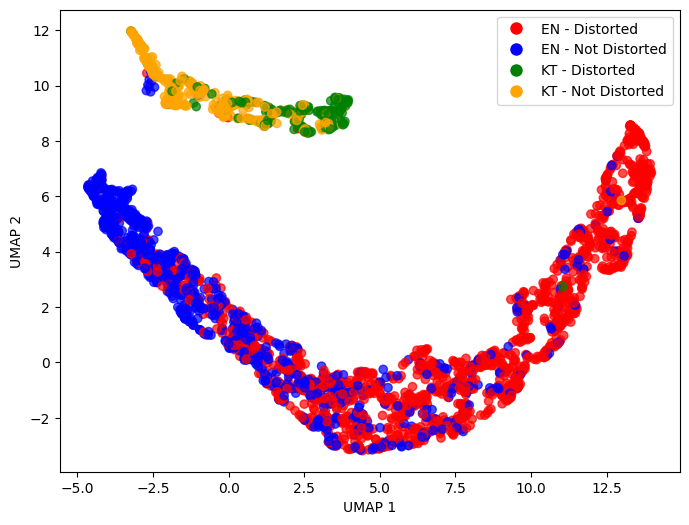}
        \caption{DCCL (EN + KT)}
        \label{fig:dccl-both}
    \end{subfigure}

    \caption{UMAP plots of the embeddings for XLM RoBERTa and DCCL. Column 1 represents embeddings of De Kindertelefoon (KT) texts, Column 2 corresponds to embeddings of English (EN) texts, and Column 3 shows both combined. Row 1 displays embeddings from XLM RoBERTa, and Row 2 from DCCL. The yellow dots represent non distorted KT posts, the green represent distorted KT posts, the blue represent non distorted EN texts and red represent distorted EN texts.}
    \label{fig:umap_embeddings}
\end{figure*}

We use UMAP \cite{mcinnes2020umapuniformmanifoldapproximation} to visualize how DCCL organizes the embedding space, projecting the embeddings of EN and KT texts into 2D.
Figure~\ref{fig:umap_embeddings} shows how the methods structure the embedding space. For XLM-RoBERTa (Figures~\ref{fig:xlmr-nl},~\ref{fig:xlmr-en}), distorted and non-distorted texts overlap heavily in the EN and KT embedding spaces, showing minimal separation. DCCL (Figures~\ref{fig:dccl-nl},~\ref{fig:dccl-en}) achieves clearer separation, suggesting it captures features relevant to cognitive distortions.
When both EN and KT posts are plotted together (Column 3), XLM-RoBERTa (Figure~\ref{fig:xlmr-both}) exhibits an obvious language divide -- EN and KT posts are clustered in clearly separated regions. In contrast, DCCL reduces this separation, indicating better cross-register alignment through distortion-specific, domain-invariant features.

\begin{thebibliography}{35}
\providecommand{\natexlab}[1]{#1}

\bibitem[{Beck(1970)}]{BECK1970184}
Aaron~T. Beck. 1970.
\newblock \href {https://doi.org/10.1016/S0005-7894(70)80030-2} {Cognitive
  therapy: Nature and relation to behavior therapy}.
\newblock \emph{Behavior Therapy}, 1(2):184--200.

\bibitem[{Bhattacharjee et~al.(2023)Bhattacharjee, Kumarage, Moraffah, and
  Liu}]{bhattacharjee-etal-2023-conda}
Amrita Bhattacharjee, Tharindu Kumarage, Raha Moraffah, and Huan Liu. 2023.
\newblock \href {https://doi.org/10.18653/v1/2023.ijcnlp-main.40} {{C}on{DA}:
  Contrastive domain adaptation for {AI}-generated text detection}.
\newblock In \emph{Proceedings of the 13th International Joint Conference on
  Natural Language Processing and the 3rd Conference of the Asia-Pacific
  Chapter of the Association for Computational Linguistics (Volume 1: Long
  Papers)}, pages 598--610, Nusa Dua, Bali. Association for Computational
  Linguistics.

\bibitem[{Chancellor and De~Choudhury(2020)}]{Chancellor2020-xe}
Stevie Chancellor and Munmun De~Choudhury. 2020.
\newblock \href {https://rdcu.be/exNFb} {Methods in predictive techniques for
  mental health status on social media: a critical review}.
\newblock \emph{NPJ Digital Medicine}, 3(1):43.

\bibitem[{Chen et~al.(2023)Chen, Lu, and Wang}]{Chen2023}
Zhiyu Chen, Yujie Lu, and William Wang. 2023.
\newblock \href {https://doi.org/10.18653/v1/2023.findings-emnlp.284}
  {Empowering psychotherapy with large language models: Cognitive distortion
  detection through diagnosis of thought prompting}.
\newblock In \emph{Findings of the Association for Computational Linguistics:
  EMNLP 2023}, pages 4295--4304, Singapore. Association for Computational
  Linguistics.

\bibitem[{Clark and Egan(2015)}]{socratic}
Gavin Clark and Sarah Egan. 2015.
\newblock \href {https://doi.org/10.1007/s10608-015-9707-3} {The socratic
  method in cognitive behavioural therapy: A narrative review}.
\newblock \emph{Cognitive Therapy and Research}, pages 1--17.

\bibitem[{Conneau et~al.(2020)Conneau, Khandelwal, Goyal, Chaudhary, Wenzek,
  Guzm{\'a}n, Grave, Ott, Zettlemoyer, and
  Stoyanov}]{conneau2020unsupervisedcrosslingualrepresentationlearning}
Alexis Conneau, Kartikay Khandelwal, Naman Goyal, Vishrav Chaudhary, Guillaume
  Wenzek, Francisco Guzm{\'a}n, Edouard Grave, Myle Ott, Luke Zettlemoyer, and
  Veselin Stoyanov. 2020.
\newblock \href {https://doi.org/10.18653/v1/2020.acl-main.747} {Unsupervised
  cross-lingual representation learning at scale}.
\newblock In \emph{Proceedings of the 58th Annual Meeting of the Association
  for Computational Linguistics}, pages 8440--8451, Online. Association for
  Computational Linguistics.

\bibitem[{Curtiss et~al.(2021)Curtiss, Levine, Ander, and Baker}]{cbtfirstline}
Joshua~E. Curtiss, Daniella~S. Levine, Ilana Ander, and Amanda~W. Baker. 2021.
\newblock \href {https://doi.org/10.1176/appi.focus.20200045}
  {Cognitive-behavioral treatments for anxiety and stress-related disorders}.
\newblock \emph{FOCUS}, 19(2):184–189.

\bibitem[{Daniel~Han and team(2023)}]{unsloth}
Michael~Han Daniel~Han and Unsloth team. 2023.
\newblock \href {http://github.com/unslothai/unsloth} {Unsloth}.

\bibitem[{David et~al.(2018)David, Cristea, and Hofmann}]{cbtgoldstandard}
Daniel David, Ioana Cristea, and Stefan~G. Hofmann. 2018.
\newblock \href {https://doi.org/10.3389/fpsyt.2018.00004} {Why cognitive
  behavioral therapy is the current gold standard of psychotherapy}.
\newblock \emph{Frontiers in Psychiatry}, 9.

\bibitem[{Du et~al.(2020)Du, Sun, Wang, Qi, and
  Liao}]{du-etal-2020-adversarial}
Chunning Du, Haifeng Sun, Jingyu Wang, Qi~Qi, and Jianxin Liao. 2020.
\newblock \href {https://doi.org/10.18653/v1/2020.acl-main.370} {Adversarial
  and domain-aware {BERT} for cross-domain sentiment analysis}.
\newblock In \emph{Proceedings of the 58th Annual Meeting of the Association
  for Computational Linguistics}, pages 4019--4028, Online. Association for
  Computational Linguistics.

\bibitem[{Dunn(1961)}]{Dunn01031961}
Olive~Jean Dunn. 1961.
\newblock \href {https://doi.org/10.1080/01621459.1961.10482090} {Multiple
  comparisons among means}.
\newblock \emph{Journal of the American Statistical Association},
  56(293):52--64.

\bibitem[{Fast et~al.(2016)Fast, Chen, and Bernstein}]{Fast_2016}
Ethan Fast, Binbin Chen, and Michael~S. Bernstein. 2016.
\newblock \href {https://doi.org/10.1145/2858036.2858535} {Empath:
  Understanding topic signals in large-scale text}.
\newblock In \emph{Proceedings of the 2016 CHI Conference on Human Factors in
  Computing Systems}, CHI '16, page 4647–4657, New York, NY, USA. Association
  for Computing Machinery.

\bibitem[{Gao et~al.(2021)Gao, Yao, and
  Chen}]{gao2022simcsesimplecontrastivelearning}
Tianyu Gao, Xingcheng Yao, and Danqi Chen. 2021.
\newblock \href {https://doi.org/10.18653/v1/2021.emnlp-main.552} {{S}im{CSE}:
  Simple contrastive learning of sentence embeddings}.
\newblock In \emph{Proceedings of the 2021 Conference on Empirical Methods in
  Natural Language Processing}, pages 6894--6910, Online and Punta Cana,
  Dominican Republic. Association for Computational Linguistics.

\bibitem[{Houlsby et~al.(2019)Houlsby, Giurgiu, Jastrzebski, Morrone,
  De~Laroussilhe, Gesmundo, Attariyan, and Gelly}]{houlsby2019parameter}
Neil Houlsby, Andrei Giurgiu, Stanislaw Jastrzebski, Bruna Morrone, Quentin
  De~Laroussilhe, Andrea Gesmundo, Mona Attariyan, and Sylvain Gelly. 2019.
\newblock \href {https://arxiv.org/abs/1902.00751} {Parameter-efficient
  transfer learning for nlp}.
\newblock In \emph{International conference on machine learning}, pages
  2790--2799. PMLR.

\bibitem[{Jiang et~al.(2024)Jiang, Yu, Zhao, Li, Song, Qi, Zhai, Luo, Wang, Fu,
  and Yang}]{jiang2024aienhancedcognitivebehavioraltherapy}
Meng Jiang, Yi~Jing Yu, Qing Zhao, Jianqiang Li, Changwei Song, Hongzhi Qi, Wei
  Zhai, Dan Luo, Xiaoqin Wang, Guanghui Fu, and Bing~Xiang Yang. 2024.
\newblock \href {https://arxiv.org/abs/2404.11449} {Ai-enhanced cognitive
  behavioral therapy: Deep learning and large language models for extracting
  cognitive pathways from social media texts}.
\newblock \emph{Preprint}, arXiv:2404.11449.

\bibitem[{Lim et~al.(2024)Lim, Kim, Choi, Sohn, and Kim}]{Lim2024}
Sehee Lim, Yejin Kim, Chi-Hyun Choi, Jy-yong Sohn, and Byung-Hoon Kim. 2024.
\newblock \href {https://doi.org/10.18653/v1/2024.clinicalnlp-1.25} {{ERD}: A
  framework for improving {LLM} reasoning for cognitive distortion
  classification}.
\newblock In \emph{Proceedings of the 6th Clinical Natural Language Processing
  Workshop}, pages 292--300, Mexico City, Mexico. Association for Computational
  Linguistics.

\bibitem[{Long et~al.(2022)Long, Luo, Wang, and Pan}]{long-etal-2022-domain}
Quanyu Long, Tianze Luo, Wenya Wang, and Sinno Pan. 2022.
\newblock \href {https://doi.org/10.18653/v1/2022.naacl-main.217} {Domain
  confused contrastive learning for unsupervised domain adaptation}.
\newblock In \emph{Proceedings of the 2022 Conference of the North American
  Chapter of the Association for Computational Linguistics: Human Language
  Technologies}, pages 2982--2995, Seattle, United States. Association for
  Computational Linguistics.

\bibitem[{Lu et~al.(2023)Lu, Huang, Zhao, Tian, Liu, and
  Li}]{lu-etal-2023-damstf}
Menglong Lu, Zhen Huang, Yunxiang Zhao, Zhiliang Tian, Yang Liu, and Dongsheng
  Li. 2023.
\newblock \href {https://doi.org/10.18653/v1/2023.acl-long.92} {{D}a{MSTF}:
  Domain adversarial learning enhanced meta self-training for domain
  adaptation}.
\newblock In \emph{Proceedings of the 61st Annual Meeting of the Association
  for Computational Linguistics (Volume 1: Long Papers)}, pages 1650--1668,
  Toronto, Canada. Association for Computational Linguistics.

\bibitem[{Luo et~al.(2022)Luo, Guo, Liu, and Zhang}]{luo-etal-2022-mere}
Yun Luo, Fang Guo, Zihan Liu, and Yue Zhang. 2022.
\newblock \href {https://aclanthology.org/2022.coling-1.620/} {Mere contrastive
  learning for cross-domain sentiment analysis}.
\newblock In \emph{Proceedings of the 29th International Conference on
  Computational Linguistics}, pages 7099--7111, Gyeongju, Republic of Korea.
  International Committee on Computational Linguistics.

\bibitem[{McInnes et~al.(2020)McInnes, Healy, and
  Melville}]{mcinnes2020umapuniformmanifoldapproximation}
Leland McInnes, John Healy, and James Melville. 2020.
\newblock \href {https://arxiv.org/abs/1802.03426} {Umap: Uniform manifold
  approximation and projection for dimension reduction}.
\newblock \emph{Preprint}, arXiv:1802.03426.

\bibitem[{McNemar(1947)}]{McNemar1947}
Quinn McNemar. 1947.
\newblock \href {https://doi.org/10.1007/bf02295996} {Note on the sampling
  error of the difference between correlated proportions or percentages}.
\newblock \emph{Psychometrika}, 12(2):153--157.

\bibitem[{Nazarova(2023)}]{Nazarova2023}
Deniz Nazarova. 2023.
\newblock \href {https://doi.org/10.1007/978-3-031-47454-5_16} {Application of
  artificial intelligence in mental healthcare: Generative pre-trained
  transformer 3 (gpt-3) and cognitive distortions}.
\newblock In \emph{Lecture Notes in Networks and Systems}, volume 813 LNNS,
  pages 204--219. Springer Science and Business Media Deutschland GmbH.

\bibitem[{Overholser and Beale(2023)}]{Overholser03102023}
James Overholser and Eleanor Beale. 2023.
\newblock \href {https://doi.org/10.1080/10503307.2023.2183154} {The art and
  science behind socratic questioning and guided discovery: a research review}.
\newblock \emph{Psychotherapy Research}, 33(7):946--956.
\newblock PMID: 36878221.

\bibitem[{Persons et~al.(2023)Persons, Marker, and Bailey}]{PERSONS2023104338}
Jacqueline~B. Persons, Craig~D. Marker, and Emily~N. Bailey. 2023.
\newblock \href {https://doi.org/10.1016/j.brat.2023.104338} {Changes in
  affective and cognitive distortion symptoms of depression are reciprocally
  related during cognitive behavior therapy}.
\newblock \emph{Behaviour Research and Therapy}, 166:104338.

\bibitem[{Ramponi and Plank(2020)}]{ramponi-plank-2020-neural}
Alan Ramponi and Barbara Plank. 2020.
\newblock \href {https://doi.org/10.18653/v1/2020.coling-main.603} {Neural
  unsupervised domain adaptation in {NLP}{---}{A} survey}.
\newblock In \emph{Proceedings of the 28th International Conference on
  Computational Linguistics}, pages 6838--6855, Barcelona, Spain (Online).
  International Committee on Computational Linguistics.

\bibitem[{Sharma et~al.(2023)Sharma, Rushton, Lin, Wadden, Lucas, Miner,
  Nguyen, and Althoff}]{sharmacognitivereframing}
Ashish Sharma, Kevin Rushton, Inna Lin, David Wadden, Khendra Lucas, Adam
  Miner, Theresa Nguyen, and Tim Althoff. 2023.
\newblock \href {https://doi.org/10.18653/v1/2023.acl-long.555} {Cognitive
  reframing of negative thoughts through human-language model interaction}.
\newblock In \emph{Proceedings of the 61st Annual Meeting of the Association
  for Computational Linguistics (Volume 1: Long Papers)}, pages 9977--10000,
  Toronto, Canada. Association for Computational Linguistics.

\bibitem[{Shreevastava and Foltz(2021)}]{Shreevastava2021}
Sagarika Shreevastava and Peter Foltz. 2021.
\newblock \href {https://doi.org/10.18653/v1/2021.clpsych-1.17} {Detecting
  cognitive distortions from patient-therapist interactions}.
\newblock In \emph{Proceedings of the Seventh Workshop on Computational
  Linguistics and Clinical Psychology: Improving Access}, pages 151--158,
  Online. Association for Computational Linguistics.

\bibitem[{Simms et~al.(2017)Simms, Ramstedt, Rich, Richards, Martinez, and
  Giraud-Carrier}]{8031202}
T.~Simms, C.~Ramstedt, M.~Rich, M.~Richards, T.~Martinez, and
  C.~Giraud-Carrier. 2017.
\newblock \href {https://doi.org/10.1109/ICHI.2017.39} {Detecting cognitive
  distortions through machine learning text analytics}.
\newblock In \emph{2017 IEEE International Conference on Healthcare Informatics
  (ICHI)}, pages 508--512.

\bibitem[{Touvron et~al.(2023)Touvron, Lavril, Izacard, Martinet, Lachaux,
  Lacroix, Rozière, Goyal, Hambro, Azhar, Rodriguez, Joulin, Grave, and
  Lample}]{touvron2023llamaopenefficientfoundation}
Hugo Touvron, Thibaut Lavril, Gautier Izacard, Xavier Martinet, Marie-Anne
  Lachaux, Timothée Lacroix, Baptiste Rozière, Naman Goyal, Eric Hambro,
  Faisal Azhar, Aurelien Rodriguez, Armand Joulin, Edouard Grave, and Guillaume
  Lample. 2023.
\newblock \href {https://arxiv.org/abs/2302.13971} {Llama: Open and efficient
  foundation language models}.
\newblock \emph{Preprint}, arXiv:2302.13971.

\bibitem[{Wang and Wu(2024)}]{wang2024stochasticadversarialnetworksmultidomain}
Xu~Wang and Yuan Wu. 2024.
\newblock \href {https://arxiv.org/abs/2406.00044} {Stochastic adversarial
  networks for multi-domain text classification}.
\newblock \emph{Preprint}, arXiv:2406.00044.

\bibitem[{Wolf et~al.(2020)Wolf, Debut, Sanh, Chaumond, Delangue, Moi, Cistac,
  Rault, Louf, Funtowicz, Davison, Shleifer, von Platen, Ma, Jernite, Plu, Xu,
  Le~Scao, Gugger, Drame, Lhoest, and Rush}]{wolf2019huggingface}
Thomas Wolf, Lysandre Debut, Victor Sanh, Julien Chaumond, Clement Delangue,
  Anthony Moi, Pierric Cistac, Tim Rault, Remi Louf, Morgan Funtowicz, Joe
  Davison, Sam Shleifer, Patrick von Platen, Clara Ma, Yacine Jernite, Julien
  Plu, Canwen Xu, Teven Le~Scao, Sylvain Gugger, Mariama Drame, Quentin Lhoest,
  and Alexander Rush. 2020.
\newblock \href {https://doi.org/10.18653/v1/2020.emnlp-demos.6} {Transformers:
  State-of-the-art natural language processing}.
\newblock In \emph{Proceedings of the 2020 Conference on Empirical Methods in
  Natural Language Processing: System Demonstrations}, pages 38--45, Online.
  Association for Computational Linguistics.

\bibitem[{{World Health Organization}(2024)}]{mentalillnesswho}
{World Health Organization}. 2024.
\newblock \href
  {https://www.who.int/news-room/fact-sheets/detail/adolescent-mental-health}
  {Mental health of adolescents}.
\newblock Accessed: 01-05-2025.

\bibitem[{Xu et~al.(2023)Xu, Wu, Yang, and
  Dai}]{xu2023foalfinegrainedcontrastivelearning}
Ting Xu, Zhen Wu, Huiyun Yang, and Xinyu Dai. 2023.
\newblock \href {https://arxiv.org/abs/2311.10373} {Foal: Fine-grained
  contrastive learning for cross-domain aspect sentiment triplet extraction}.
\newblock \emph{Preprint}, arXiv:2311.10373.

\bibitem[{Zhan et~al.(2024)Zhan, Zheng, Lee, Suh, Li, and
  Ong}]{zhan2024largelanguagemodelscapable}
Hongli Zhan, Allen Zheng, Yoon~Kyung Lee, Jina Suh, Junyi~Jessy Li, and
  Desmond~C. Ong. 2024.
\newblock \href {https://arxiv.org/abs/2404.01288} {Large language models are
  capable of offering cognitive reappraisal, if guided}.
\newblock \emph{Preprint}, arXiv:2404.01288.

\bibitem[{Zhou et~al.(2020)Zhou, Tian, Wang, Wu, Xiao, and
  He}]{zhou-etal-2020-sentix}
Jie Zhou, Junfeng Tian, Rui Wang, Yuanbin Wu, Wenming Xiao, and Liang He. 2020.
\newblock \href {https://doi.org/10.18653/v1/2020.coling-main.49} {{S}enti{X}:
  A sentiment-aware pre-trained model for cross-domain sentiment analysis}.
\newblock In \emph{Proceedings of the 28th International Conference on
  Computational Linguistics}, pages 568--579, Barcelona, Spain (Online).
  International Committee on Computational Linguistics.

\end{thebibliography}
\end{document}